\documentclass[runningheads]{llncs}
\usepackage[T1]{fontenc}
\usepackage{graphicx}
\usepackage{amsmath}  
\usepackage{amssymb}
\usepackage{booktabs}
\begin{document}
\title{MSPLoRA: A Multi-Scale Pyramid Low-Rank Adaptation for Efficient Model Fine-Tuning}
%
%

\author{
Jiancheng Zhao\inst{1}\textsuperscript{*}\orcidID{0009-0003-3055-5584} \and
Xingda Yu\inst{1}\textsuperscript{*}\orcidID{0009-0002-0167-1159} \and
Zhen Yang\inst{1}\textsuperscript{$\dagger$}\orcidID{0000-0003-0670-4538}
}
\authorrunning{Zhao et al.}
%
\institute{Shandong University, Qingdao, China
\email{\{202200120166,202200120014,zhenyang\}@sdu.edu.cn}\\
}
\maketitle
\begingroup
\renewcommand\thefootnote{}\footnotetext{\textsuperscript{*} Equal contribution.}
\renewcommand\thefootnote{}\footnotetext{\textsuperscript{$\dagger$} Corresponding author.}
\endgroup

\begin{abstract}
Parameter-Efficient Fine-Tuning (PEFT) has become an essential approach for adapting large-scale pre-trained models while reducing computational costs. Among PEFT methods, LoRA significantly reduces trainable parameters by decomposing weight updates into low-rank matrices. However, traditional LoRA applies a fixed rank across all layers, failing to account for the varying complexity of hierarchical information, which leads to inefficient adaptation and redundancy. To address this, we propose MSPLoRA (Multi-Scale Pyramid LoRA), which introduces Global Shared LoRA, Mid-Level Shared LoRA, and Layer-Specific LoRA to capture global patterns, mid-level features, and fine-grained information, respectively. This hierarchical structure reduces inter-layer redundancy while maintaining strong adaptation capability. Experiments on various NLP tasks demonstrate that MSPLoRA achieves more efficient adaptation and better performance while significantly reducing the number of trainable parameters. Furthermore, additional analyses based on Singular Value Decomposition validate its information decoupling ability, highlighting MSPLoRA as a scalable and effective optimization strategy for parameter-efficient fine-tuning in large language models. Our code is available at https://github.com/Oblivioniss/MSPLoRA.

\end{abstract}
\section{Introduction}
Recently, with the rapid advancement of large-scale pre-trained models such as GPT-4~\cite{gpt}, the pre-training–fine-tuning paradigm has become the dominant approach, demonstrating remarkable capabilities and superior performance across various tasks, including image processing~\cite{vlm} and code translation~\cite{code}. Fine-tuning serves as a critical step in aligning pre-trained models, which have acquired extensive knowledge, to specific downstream tasks~\cite{ft}. Numerous studies have validated its significance and effectiveness.

However, as the parameter scale of modern large language models continues to grow, traditional full-parameter fine-tuning is becoming increasingly impractical due to computational and storage constraints. To address this challenge, various PEFT methods have been proposed. Among them, LoRA~\cite{lora(adapter2)} leverages the low-rank nature of weight updates by decomposing certain weight matrices into the product of low-rank matrices. This significantly reduces the number of trainable parameters without introducing additional inference latency, thereby mitigating computational and storage overhead. Nonetheless, traditional LoRA employs a fixed low-rank decomposition across all layers, treating different layers uniformly and failing to account for their distinct information processing requirements. While approaches such as AdaLoRA~\cite{adalora(onlyqv)} attempt to dynamically adjust the LoRA rank for different layers, existing LoRA-based methods still lack the capability to model multi-scale information, leading to the entanglement of global patterns, local features, and fine-grained information within a single low-rank space. This limitation may hinder fine-tuning efficiency and generalization. Consequently, an important question arises:

\begin{quote}
    {\textit{How can we decouple global patterns, mid-level features, and fine-grained information without increasing computational overhead, thereby enhancing the adaptation of pre-trained models?}}
\end{quote}

To address this issue, Lily~\cite{lily} introduces a globally shared projection matrix while keeping each down-projection matrix independent and employs a routing network to dynamically match them. However, this approach does not explicitly disentangle shared information across hierarchical levels, and its reliance on a mixture-of-experts structure leaves further room for parameter reduction. In contrast, a hierarchical pyramid structure naturally accommodates multi-scale information modeling, allowing different levels to be adapted with distinct strategies so that global patterns, mid-level features, and fine-grained information can be separately captured and modeled.

\begin{figure}[t]
\includegraphics[width=\textwidth]{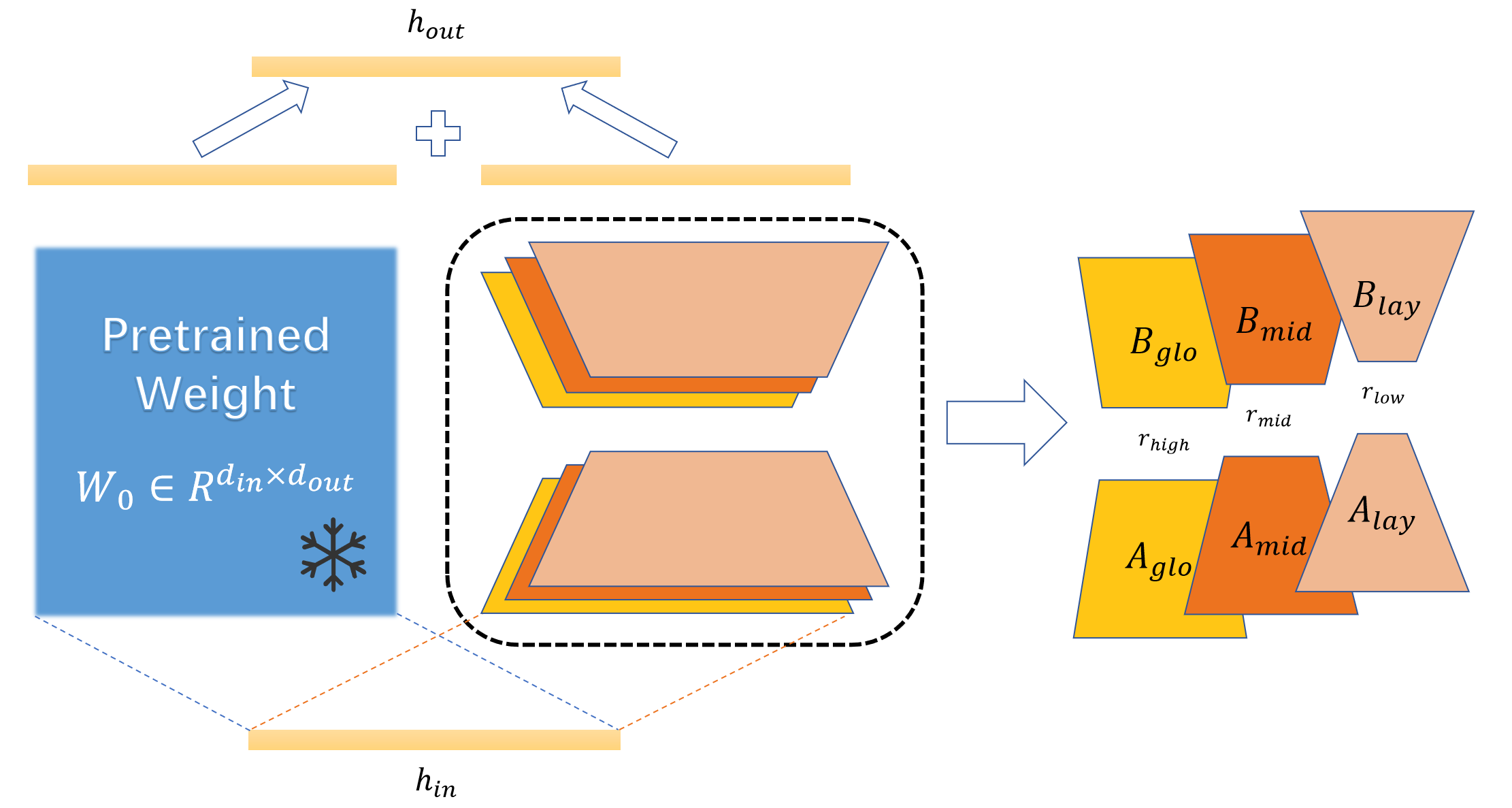}
\caption{The multi-scale pyramid structure of MSPLoRA is illustrated, where the pretrained weights remain frozen, and the LoRA components are divided into global shared LoRA, mid-level shared LoRA, and layer-specific LoRA, which respectively learn global patterns, mid-level features, and fine-grained information. Meanwhile, the structure on the right indicates that the rank of LoRA components gradually decreases from large-scale to small-scale, enabling more targeted information modeling at different hierarchical levels while reducing parameter redundancy and improving fine-tuning efficiency.} \label{fig1}
\end{figure}

In this work, we propose MSPLoRA (Multi-Scale Pyramid LoRA), a novel framework that introduces a hierarchical low-rank adaptation mechanism. As shown in Fig 1, MSPLoRA constructs a multi-scale LoRA structure to decouple global, mid-level, and fine-grained information. Specifically, at the global scale, given the richness and complexity of global information, we employ a globally shared LoRA component with a higher rank to learn overarching patterns. At the mid-level scale, we introduce a locally shared LoRA component with a moderate rank, capturing shared features within subsets of layers. At the layer-specific scale, considering the unique characteristics of each layer, we employ independent LoRA components with the lowest rank, enabling the adaptation of fine-grained information at each layer.
By adopting this pyramidal structure, MSPLoRA significantly reduces the number of trainable parameters compared to standard LoRA, while simultaneously improving parameter efficiency and adaptation quality. To evaluate the effectiveness of MSPLoRA, we conduct extensive experiments across different models and tasks, demonstrating its superior performance and efficiency.For natural language understanding (NLU) tasks, we fine-tune RoBERTa-base on the GLUE benchmark and evaluate its performance.For instruction-following tasks, we fine-tune LLaMA2-7B on the Alpaca dataset and assess its problem-solving capability using INSTRUCTEVAL. The experimental results demonstrate that MSPLoRA outperforms LoRA and its variants in terms of both parameter efficiency and overall performance across multiple tasks and models. Our contributions are as follows:

\renewcommand{\labelitemi}{$\bullet$}

\begin{itemize}
    \item We propose MSPLoRA, a multi-scale pyramid LoRA framework that decomposes information across different scales, leading to more efficient parameter utilization and improved adaptation to diverse tasks.
    \item We conduct comprehensive experiments across multiple models and tasks, showing that MSPLoRA consistently outperforms LoRA across different settings.
    \item We further analyze the rank assignment of different LoRA components in MSPLoRA and validate its effectiveness in reducing inter-layer redundancy using Singular Value Decomposition (SVD), demonstrating that MSPLoRA enhances model adaptation by better disentangling multi-scale information.
\end{itemize}

\section{Related Work}
\subsection{Parameter Efficient Fine-Tuning}
In recent years, with the continuous growth of parameter sizes in large-scale pre-trained models such as BERT~\cite{bert}, LLaMA~\cite{llama}, and T5~\cite{t5}, traditional full-parameter fine-tuning has faced numerous challenges, including high computational overhead and significant storage costs. More critically, studies have shown that full fine-tuning may lead to catastrophic forgetting of knowledge acquired during the pre-training phase. To address these issues, Parameter-Efficient Fine-Tuning (PEFT) has been proposed, aiming to align pre-trained models with downstream tasks by modifying only a small subset of parameters. This approach not only reduces storage and computational costs but also helps retain most of the model’s original knowledge. Currently, mainstream PEFT methods can be categorized into three types. Adapter-based approaches~\cite{adapter1,lora(adapter2)} freeze the original model parameters and introduce lightweight adapter modules to enable fine-tuning through additional trainable structures. Prompt-based methods, including Prefix-Tuning~\cite{prefixtuning} and P-Tuning~\cite{ptuning1,ptuning2}, modify the model's input or embedding vectors by adding trainable prefixes to better align with downstream tasks. Low-Rank Adaptation (LoRA)~\cite{lora(adapter2)} methods leverage low-rank decomposition by inserting low-rank matrices at specific positions and updating only these matrices during fine-tuning to adjust the model weights. As an emerging research direction in recent years, PEFT has provided a powerful tool for efficiently adapting large-scale pre-trained models and has been widely applied across various domains. ~\cite{codepeft,clinical,financial}

\subsection{Low-Rank Adaption}
LoRA (Low-Rank Adaptation)~\cite{lora(adapter2)} is a Parameter-Efficient Fine-Tuning (PEFT) method that introduces a low-rank downsampling matrix \( A \) and a low-rank upsampling matrix \( B \) at specific locations in the model (e.g., the attention layers in transformer models). During fine-tuning, the pre-trained model weights remain frozen, and only these low-rank matrices are updated. This process can be represented as \(\Delta W = AB\) where \( \Delta W \) denotes the weight update relative to the pre-trained model. Compared to full-parameter fine-tuning, LoRA significantly reduces the number of trainable parameters while maintaining model performance, and it introduces no additional inference latency. However, traditional LoRA still leaves room for further parameter reduction and employs a fixed-rank decomposition strategy, which fails to flexibly adapt to different hierarchical information needs. This limitation may lead to overfitting or underfitting in certain layers, negatively impacting fine-tuning performance. To address these limitations, researchers have proposed various LoRA variants to enhance its adaptability. AdaLoRA~\cite{adalora(onlyqv)} dynamically allocates different LoRA ranks across layers based on the singular value decomposition of weight updates, assigning higher ranks to more important layers while using lower ranks for less important layers, thereby improving parameter efficiency and flexibility. ASLoRA~\cite{aslora} employs cross-layer parameter sharing, where the low-rank matrix \( A \) is shared across all layers, and matrix \( B \) is adaptively merged to reduce redundant parameters, improving parameter efficiency while maintaining or even enhancing fine-tuning performance. Q-LoRA~\cite{qlora} integrates 4-bit quantization with LoRA, enabling low-storage-cost fine-tuning while maintaining strong performance. Although these approaches have improved LoRA’s efficiency and adaptation capabilities, they do not explicitly model the hierarchical information needs of different layers. As a result, global patterns, mid-level structures, and fine-grained features remain entangled within the same low-rank space, potentially limiting fine-tuning effectiveness. To address this issue, we propose MSPLoRA, which introduces a multi-scale pyramid decomposition structure that explicitly models global patterns, mid-level features, and fine-grained information at different levels, thereby enhancing parameter efficiency and adaptation capability in fine-tuning.

\subsection{Hierarchical Pyramid Structure}
The Hierarchical Pyramid Structure, as a multi-scale modeling approach, has been widely applied in computer vision, natural language processing, and representation learning. SPP-Net (Spatial Pyramid Pooling Network)~\cite{sppnet} extracts features at different scales through a pyramid pooling mechanism, allowing CNNs to handle variable-sized inputs without fixed cropping, significantly improving model flexibility. FPN~\cite{fpn} enhances object detection capabilities through top-down feature fusion, while HRNet~\cite{hrnet} strengthens high-resolution feature representation by facilitating multi-scale information interaction. Additionally, Deformable DETR~\cite{detr} combines pyramid structures with deformable attention to optimize small object detection. These methods all leverage pyramid structures to enhance the representation of multi-scale information. However, despite the success of pyramid structures in fields such as computer vision, their application in PEFT remains relatively limited. To address this, we propose MSPLoRA, which draws inspiration from pyramid-based modeling and introduces a multi-scale decomposition mechanism into the LoRA adaptation framework. By enabling different LoRA components to model global patterns, mid-level features, and fine-grained information separately, MSPLoRA improves parameter efficiency and enhances fine-tuning quality.

\section{Method}
\subsection{Preliminaries on Low-Rank Adapter}
LoRA approximates weight updates through low-rank decomposition, expressed as:

\begin{equation}
\Delta W = AB
\end{equation}

where \( A \) and \( B \) are two trainable low-rank matrices, and \( r \) represents the rank of LoRA. Compared to full fine-tuning, LoRA only optimizes the parameters of \( A \) and \( B \) while keeping the original weight matrix \( W \) frozen during fine-tuning, thereby reducing both the computational and storage overhead required for training.  

To ensure that LoRA does not interfere with the model's original behavior during the early stages of fine-tuning, it is initialized as follows:

\begin{equation}
A \sim \mathcal{N}(0, \sigma^2), \quad B = 0
\end{equation}

where the elements of \( A \) are randomly initialized from a Gaussian distribution with mean 0 and variance \( \sigma^2 \), while \( B \) is initialized as an all-zero matrix.

\subsection{Multi-Scale Pyramid Low-Rank Adaptation}

\begin{figure}[t]
\includegraphics[width=\textwidth]{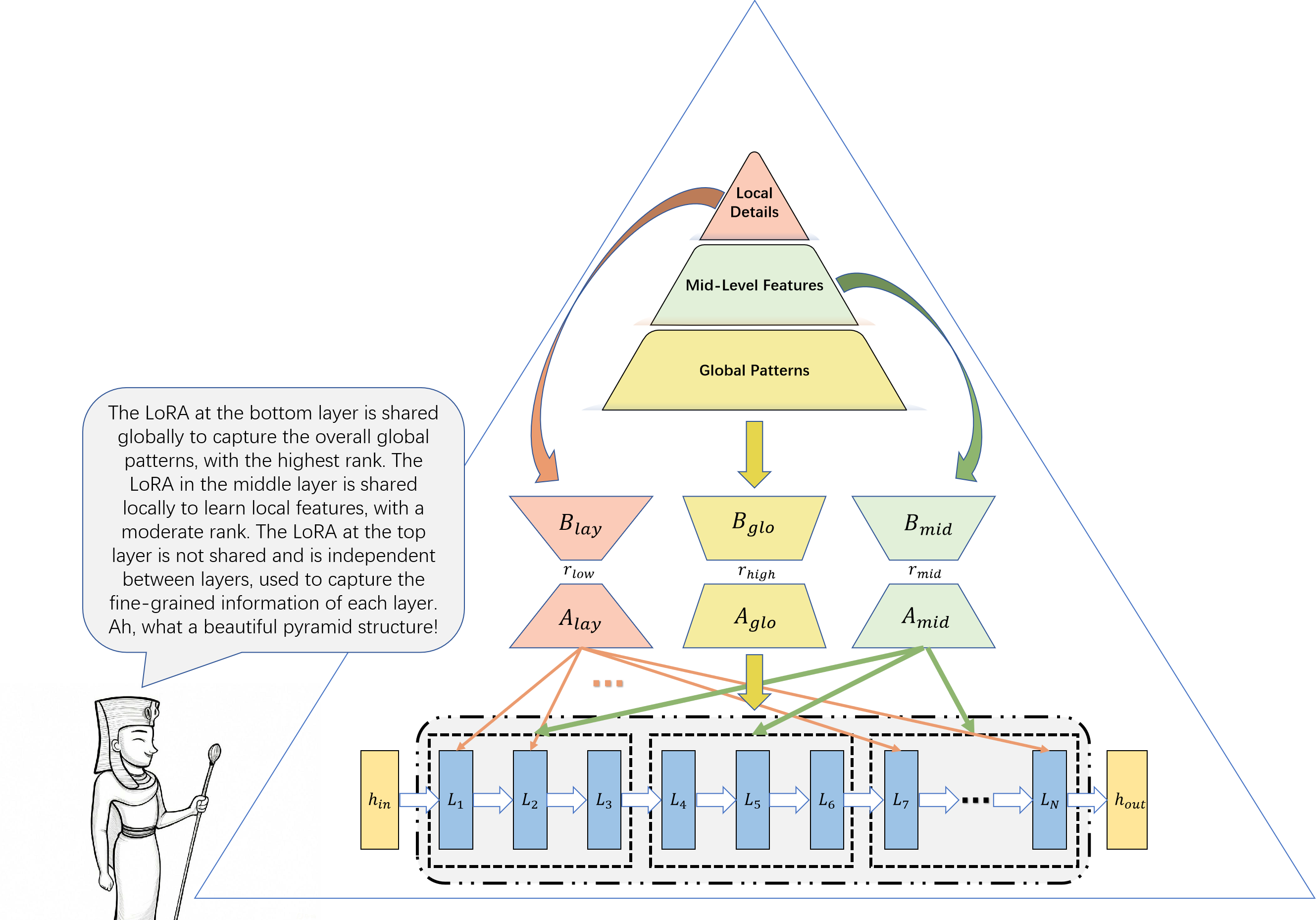}
\caption{Illustration of the multi-scale pyramid structure of MSPLoRA and its application to the Transformer hierarchy.} \label{fig2}
\end{figure}

As shown in Fig 2, In MSPLoRA, we introduce a multi-scale pyramid structure to achieve a hierarchical decomposition of LoRA components, allowing different scales of LoRA components to learn global, mid-level, and fine-grained information separately. Given the weight matrix \( W \) of a certain layer in the pre-trained model, MSPLoRA introduces three LoRA components, namely global shared LoRA, mid-level shared LoRA, and layer-specific LoRA, so that the weight update can be expressed as:

\begin{equation}
\Delta W = \Delta W_{\text{global}} + \Delta W_{\text{mid-level}} + \Delta W_{\text{layer-specific}}
\end{equation}

where \( \Delta W_{\text{global}} \) represents the contribution of the global LoRA update, \( \Delta W_{\text{mid-level}} \) represents the contribution of the mid-level LoRA update, and \( \Delta W_{\text{layer-specific}} \) represents the contribution of the layer-specific LoRA update. Each weight update component can be further formulated as:

\begin{equation}
\Delta W_{\text{global}} = A_{\text{g}} B_{\text{g}}, \quad
\Delta W_{\text{mid-level}} = A_{\text{m}} B_{\text{m}}, \quad
\Delta W_{\text{layer-specific}} = A_{\text{l}} B_{\text{l}}
\end{equation}

where \( A_{\text{g}} \) and \( B_{\text{g}} \) denote the downsampling and upsampling matrices of the global LoRA component, which are shared across all layers in the model. Considering that global information is usually more complex, we adopt a higher rank for this component. Similarly, \( A_{\text{m}} \) and \( B_{\text{m}} \) represent the downsampling and upsampling matrices of the mid-level LoRA component, which is shared among certain layers and assigned a moderate rank. Finally, \( A_{\text{l}} \) and \( B_{\text{l}} \) correspond to the downsampling and upsampling matrices of the layer-specific LoRA component, which is independently applied at each layer. Since layer-specific information is typically more sparse, we assign it a lower rank.

\subsection{Advantages Analysis}
\subsubsection{MSPLoRA Reduces Redundancy through Pyramid Decomposition.} Currently, the Transformer architecture is widely adopted in large language models, where the core structure consists of stacked encoder and decoder layers. This repetitive design naturally introduces a certain degree of redundancy. Studies by Dalvi et al.~\cite{dalvi} have shown that large-scale pre-trained Transformer models contain a significant number of redundant neurons, with the highest redundancy observed in adjacent layers. Additionally, Bhojanapalli et al.~\cite{bhojanapalli} found that in pre-trained Transformer-based models such as BERT and ViT, attention scores between neighboring layers exhibit high redundancy, indicating that information across layers is not independent but highly correlated. Therefore, efficiently utilizing inter-layer shared information while reducing redundancy has become a key challenge in optimizing Transformer-based architectures. To tackle this issue, MSPLoRA employs a multi-scale pyramid decomposition mechanism, where LoRA components at different scales learn distinct granularities of information. Unlike traditional LoRA, which applies a uniform low-rank decomposition across all layers, MSPLoRA enables global shared LoRA components to capture cross-layer common features, mid-level shared LoRA components to focus on subsets of layers, and layer-specific LoRA components to retain fine-grained information for crucial layers. This hierarchical structure effectively reduces redundancy across LoRA components at different layers, resulting in a more compact parameter space for LoRA adaptation, which enhances both fine-tuning efficiency and adaptation capability.

\subsubsection{MSPLoRA Significantly Reduces Trainable Parameters.} 
MSPLoRA Significantly Reduces Trainable Parameters. When using LoRA for fine-tuning, for an \(N\)-layer model, LoRA components are typically added to different modules at each layer. A common setup involves adding a set of low-rank matrices \( A \in \mathbb{R}^{d \times r_0} \), \( B \in \mathbb{R}^{r_0 \times d} \) to query and value projection layers (\( W_q, W_v \)), making the total number of LoRA parameters:

\begin{equation}
\text{Total Parameters}_{\text{LoRA}} = 4N r_0 d
\end{equation}

where \(d\) is the embedding dimension and \(r_0\) is the rank of LoRA.For MSPLoRA, in our experiments, we simply set the ranks of different LoRA components to decay geometrically, meaning:

\begin{equation}
r_{\text{mid}} = \frac{r_{\text{high}}}{2}, \quad r_{\text{low}} = \frac{r_{\text{mid}}}{2}
\end{equation}

where \(r_\text{high}\) is the rank of the globally shared LoRA, \(r_\text{mid}\) is the rank of the mid-level shared LoRA, and \(r_\text{low}\) is the rank of the layer-specific LoRA. Additionally, regarding the division of locally shared regions, Nadipalli et al.~\cite{nadipalli} pointed out that fine-tuned Transformer models can be divided into three functional layers: the lower layers, which primarily retain general linguistic representations; the middle layers, which serve as a transition from general to task-specific features; and the upper layers, which focus on task-specific adaptation. Based on this, we define three distinct mid-level LoRA components, each dedicated to extracting locally shared information in the lower, middle, and upper layers.

Since different models may have architectural variations, the definitions of lower, middle, and upper layers may vary. To address this, we uniformly divide the layers based on model depth. If we set \(r_{\text{high}} = r_{\text{0}}\), then the total parameter count in MSPLoRA is given by:

\begin{equation}
\text{Total Parameters}_{\text{MSPLoRA}} = 4r_0d + 3 \times \frac{4r_0d}{2} + N \times \frac{4r_0d}{4} = (10 + N) r_0 d
\end{equation}

Thus, for deep large-scale pre-trained models, MSPLoRA effectively decouples information across different scales while achieving a higher effective rank per layer, requiring only one-fourth of the parameters compared to standard LoRA.

\begin{table}[h]
\caption{Results on GLUE for natural language understanding tasks.}\label{tab1}
    \centering
    \resizebox{\textwidth}{!}{
    \renewcommand{\arraystretch}{1.2}  
    \setlength{\tabcolsep}{6pt}  
    \begin{tabular}{l|c|c|c|c|c|c|c|c}  
        \toprule
        \textbf{Method} & \# \textbf{Trainable} & \textbf{SST-2} & \textbf{MRPC} & \textbf{CoLA} & \textbf{QNLI} & \textbf{RTE} & \textbf{STS-B} & \textbf{Avg.} \\
        & \textbf{Parameters} & (Acc.) & (Acc.) & (MCC) & (Acc.) & (Acc.) & (PCC) & \\
        \midrule
        FFT   & 125M  & \underline{94.8}  & 90.2  & 63.6  & 92.8  & 78.7  & 91.2  & 85.2 \\
        \midrule  
        BitFit  & 0.1M  & 93.7  & \textbf{92.7}  & 62.0  & 91.8  & 81.5  & 90.8  & 85.4 \\
        Adpt$^{D}$  & 0.3M  & 94.2  & 88.5  & 60.8  & 93.1  & 71.5  & 89.7  & 83.0 \\
        Adpt$^{D}$  & 0.9M  & 94.7  & 88.4  & 62.6  & 93.0  & 75.9  & 90.3  & 84.2 \\
        LoRA    & 0.3M  & \textbf{95.1}  & 89.7  & 63.4  & \textbf{93.3}  & 78.4  & \textbf{91.5}  & 85.2 \\
        AdaLoRA & 0.3M  & 94.5  & 88.7  & 62.0  & 93.1  & 81.0  & 90.5  & 85.0 \\
        DyLoRA  & 0.3M  & 94.3  & 89.5  & 61.1  & 92.2  & 78.7  & 91.1  & 84.5 \\
        Lily  & 0.3M  & 95.0  & 90.2  & \textbf{66.0}  & 92.5  & 81.6  & 90.8  & \underline{86.0} \\
        \textbf{MSPLoRA}  & 0.135M  & 94.7  & \underline{90.5}  & \underline{65.6}  & \underline{93.1}  & \textbf{81.7}  & \underline{91.2}  & \textbf{86.2} \\
        \bottomrule
    \end{tabular}
    }
    \label{tab:results}
\end{table}

\section{Experiments}
To comprehensively evaluate the fine-tuning performance of MSPLoRA across different tasks, we conduct experiments in two domains: Natural Language Understanding (NLU) and Instruction Following, selecting appropriate models and datasets for fine-tuning and evaluation. For the natural language understanding task, we use RoBERTa-base as the base model and fine-tune it on the GLUE benchmark. For the instruction-following task, we use LLaMA2-7B as the base model, fine-tune it on the Alpaca dataset, and evaluate its performance on the INSTRUCTEVAL benchmark.

\subsection{Baselines}
We compare MSPLoRA with FFT(Full Fine-Tuning), LoRA, various LoRA-based improvements, and other PEFT methods.

\renewcommand{\labelitemi}{$\bullet$}

\begin{itemize}
    \item \textbf{Full Fine-Tuning} refers to updating all the parameters of a pre-trained model on a downstream task to fully adapt it to the new data distribution.
    \item \textbf{Adapter-Tuning}~\cite{adapter1} introduces lightweight adapter modules between layers of the pre-trained model. It updates only the parameters within the adapter while keeping the original model weights frozen, thereby reducing the number of trainable parameters while achieving efficient task adaptation and transfer learning. AdapterDrop~\cite{adapterd} further optimizes this approach by selectively dropping certain adapter layers during training or inference, reducing computational cost while maintaining adaptation capability and performance.
    \item \textbf{LoRA}~\cite{lora(adapter2)} introduces low-rank matrices into specific layers of the pre-trained model to update the weights, modifying only a small number of parameters.
    \item \textbf{AdaLoRA}~\cite{adalora(onlyqv)} dynamically adjusts the LoRA rank across different layers using singular value decomposition, allowing for adaptive rank assignment based on layer importance to improve parameter efficiency.
    \item \textbf{DyLoRA}~\cite{dylora} introduces dynamic rank learning, enabling LoRA ranks to adaptively adjust during training without relying on additional importance scores or predefined rules.
    \item \textbf{PISSA}~\cite{pissa} applies singular value decomposition on pre-trained weight matrices, directly optimizing principal singular values and singular vectors while freezing the residual components.
\end{itemize}

\subsection{Natural Language Understanding}
\subsubsection{Datasets and Models} We evaluate our method by fine-tuning and assessing the RoBERTa-base~\cite{roberta} model on the GLUE benchmark~\cite{glue}. GLUE (General Language Understanding Evaluation) is a benchmark designed to evaluate natural language understanding (NLU) models covering diverse tasks. These tasks comprehensively measure MSPLoRA's performance in text understanding, reasoning, and classification.

\renewcommand{\labelitemi}{$\bullet$}

\begin{itemize}
    \item \textbf{QNLI}~\cite{qnli} Adapted from the SQuAD dataset, this task requires determining whether a given passage contains the answer to a provided question.
    \item \textbf{RTE} Determines whether there is an entailment relationship between two sentences, sourced from multiple Natural Language Inference 
    \item \textbf{MRPC}~\cite{mrpc} A paraphrase identification task that assesses whether two sentences are semantically equivalent.
    \item \textbf{STS-B}~\cite{stsb} A regression task that scores the semantic similarity of sentence pairs on a scale from 0 to 5.
    \item \textbf{SST-2}~\cite{sst2} A binary sentiment classification task based on movie reviews, predicting whether the sentiment is positive or negative.
    \item \textbf{CoLA}~\cite{cola} Evaluates whether an English sentence conforms to grammatical rules, testing a model’s understanding of syntactic correctness.
\end{itemize}

\subsubsection{Implementation Details} In our experiments, we only fine-tune the \(W_q\) and \(W_v\) at each layer~\cite{adalora(onlyqv)}. Following the convention of other LoRA-based methods, we set the rank to 8 for all baseline approaches. For MSPLoRA, as described in Section 3.3, we adopt a simple rank decay strategy from large-scale to small-scale LoRA components, where the global LoRA rank is set to 8, the mid-level LoRA rank is set to 4, and the layer-specific LoRA rank is set to 2. Additionally, we define three distinct mid-level LoRA components, corresponding to shared representations in the lower, middle, and upper layers. 

\subsubsection{Results}
The performance of all methods on the GLUE dataset is reported in Table 1, where the best results are highlighted in bold and the second-best results are underlined. As shown, our method reduces the number of trainable parameters by more than half compared to LoRA, achieves the best performance on RTE, ranks second on several sub-tasks including MRPC, CoLA, QNLI, and STS-B, and obtains the highest average score across all sub-tasks. These results clearly demonstrate the effectiveness of our method on natural language understanding tasks.

\begin{table}[h]
\caption{Results on INSTRUCTEVAL for instruction following
 tasks.}\label{tab2}
    \centering
    \renewcommand{\arraystretch}{1.2}  
    \setlength{\tabcolsep}{6pt}  
    \begin{tabular}{l|c|c|c|c|c|c}  
        \toprule
        \textbf{Method} & \# \textbf{Trainable} & \textbf{MMLU} & \textbf{BBH} & \textbf{DROP} & \textbf{HEvaL} & \textbf{Avg.} \\
        & \textbf{Parameters} \\
        \midrule
        w/o FT   & -  & 45.96  & 32.04  & 31.55  & 12.20  & 30.44 \\
        FT   & 7B  & \textbf{47.30}  & 32.72  & 29.12  & 12.80  & 30.49  \\
        \midrule
        LoRA    & 33.6M  & 45.64  & 32.40  & \textbf{32.46}  & 15.09  & \underline{31.40} \\
        AdaLoRA & 33.6M  & 45.96  & \underline{32.85}  & \underline{31.94}  & 14.02  & 31.19 \\
        QLoRA  & 33.6M  & 45.40  & 32.81  & 28.97  & \underline{15.24}  & 30.61 \\
        \textbf{MSPLoRA}  & 11.0M  & \underline{46.31}  & \textbf{32.91}  & 31.92  & \textbf{17.23}  & \textbf{32.09} \\
        \bottomrule
    \end{tabular}
    \label{tab:results}
\end{table}

\subsection{Instruction Following}
\subsubsection{Datasets and Models} For the instruction-following task, we select the LLaMA2-7B~\cite{llama2} model as our backbone and fine-tune it on the Alpaca dataset~\cite{alpaca}. Alpaca is an instruction-following dataset that contains a large number of instruction-response pairs generated by GPT-3.5, widely used to evaluate large language models in terms of multi-task learning and generalization capabilities. To further quantify the model’s ability to understand and execute instructions, we conduct evaluations on the INSTRUCTEVAL benchmark~\cite{instructeval}, which covers various real-world tasks such as code generation, mathematical reasoning, and dialogue understanding.

\renewcommand{\labelitemi}{$\bullet$}

\begin{itemize}
    \item \textbf{MMLU}~\cite{mmlu} A benchmark designed to evaluate the multi-task language understanding capabilities of large language models. It covers 57 subject areas (such as mathematics, history, physics, and law) and assesses generalization and reasoning abilities through multiple-choice 
    \item \textbf{BBH}~\cite{bbh} A subset of BIG-Bench, consisting of 23 of the most challenging tasks, covering areas such as complex reasoning, mathematics, common sense reasoning, and code comprehension.
    \item \textbf{DROP}~\cite{drop}  A reading comprehension dataset focused on discrete reasoning, requiring models to extract information from passages and perform numerical calculations, counting, and sorting-based reasoning tasks.
    \item \textbf{HumanEval}~\cite{heval} A benchmark for evaluating code generation capabilities, containing various Python programming tasks where models must generate correct code based on natural language descriptions. The generated code is automatically evaluated using unit tests to assess its correctness and execution ability.
\end{itemize}

\subsubsection{Implementation Details} 
For all other methods, we configure the rank \( r \) as 64. In the case of MSPLoRA, the global-scale LoRA rank is assigned a value of 64, while the mid-level LoRA is configured with a rank of 32, and the layer-specific LoRA is allocated a rank of 16. Regarding task configurations, MMLU is evaluated using a 5-shot direct prompting approach, while BBH and DROP (dev) adopt 3-shot direct prompting, and HumanEval follows a 0-shot direct prompting strategy. During fine-tuning, we utilize the AdamW optimizer and train each model for a total of three epochs. The learning rate follows a linear decay schedule, starting at \( 3 \times 10^{-4} \), and the batch size is specified as 128. These experimental settings maintain consistency across all tasks, facilitating a thorough and unbiased performance comparison of different models.

\subsubsection{Results}
The evaluation results of all methods on INSTRUCTEVAL are presented in Table 2, where the best results are highlighted in bold and the second-best results are underlined. Our method achieves the best performance on both the BBH and HEval sub-tasks, ranks second on MMLU, and obtains the highest overall average score. Notably, it accomplishes this while using less than one-third of the parameters compared to LoRA, demonstrating outstanding performance with significantly reduced parameter overhead. These results highlight the efficient adaptation capability of MSPLoRA on instruction-following tasks, indicating its ability to capture the core information required for instruction tuning while substantially lowering computational costs.

\begin{figure}[t]
\includegraphics[width=\textwidth]{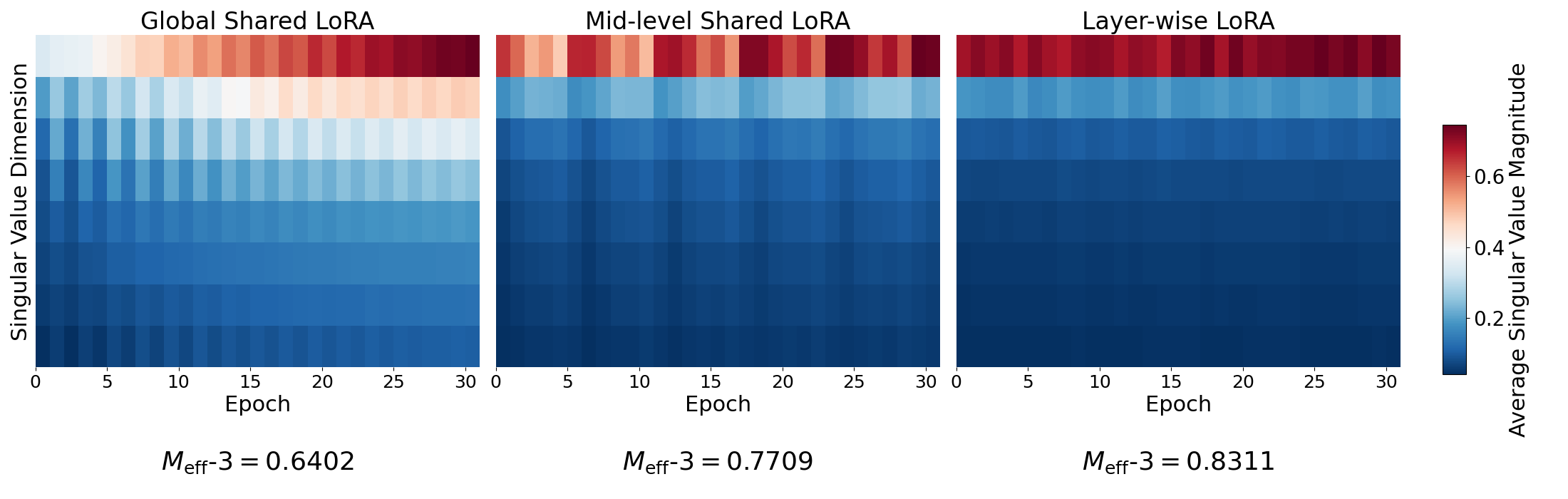}
\caption{This figure shows the SVD heatmaps of different LoRA components in MSPLoRA. The x-axis denotes training epochs, and the y-axis represents singular value dimensions, with color indicating average magnitude. The global LoRA (left) has the highest singular value intensity and active dimensions, followed by the mid-level shared LoRA (middle), while the layer-specific LoRA (right) has the lowest. This supports the effectiveness of the proposed rank settings in our multi-scale pyramid design.} \label{fig3}
\end{figure}

\section{Further Analysis}
To further explore the effectiveness of MSPLoRA, we conducted an in-depth analysis to validate the rationality of rank settings for different LoRA components and to demonstrate that MSPLoRA effectively reduces inter-layer information redundancy compared to standard LoRA.
\subsection{Analysis of Rank Setting Rationality}
As described in Section 3.3, we simply set the MSPLoRA ranks to decay geometrically from large-scale to small-scale LoRA components. To validate the effectiveness of this design, we uniformly assign high ranks to the global, mid-level shared, and layer-specific LoRA components, and fine-tune a RoBERTa-base model on three tasks—MRPC, STS-B, and CoLA—while recording the singular value decomposition of the weight update matrices during training to analyze the information complexity learned by each LoRA component. Specifically, for a given matrix, we measure its effective rank space by computing the ratio of the sum of its top k singular values to the sum of all singular values:

\begin{equation}
    M_{\text{eff}}\text{-}k = \frac{\sum_{i=1}^{k} \sigma_i}{\sum_{j=1}^{M} \sigma_j}
\end{equation}

where \( M \) represents the original rank of the matrix, and \( \sigma_i \) denotes the i-th singular value, sorted in descending order. The metric \( M_{\text{eff}} \) indicates the effective rank space of the matrix. A higher \( M_{\text{eff}} \) value suggests that the majority of the matrix's information is concentrated in the top singular values, meaning the matrix has strong low-rank characteristics and lower information complexity. Conversely, a lower \( M_{\text{eff}} \) value suggests that information is more evenly distributed across multiple singular value directions, indicating higher complexity.

We record the top-8 singular values (ranked in descending order) of the global, mid-level, and layer-specific LoRA components, and visualize their averaged heatmaps to examine the information complexity at different scales. Additionally, we compute the effective rank space to quantitatively assess their expressive capacity. As shown in Fig 3, the experimental results demonstrate that the global shared LoRA exhibits the highest effective rank, indicating a broader distribution of singular values and higher information complexity. This supports the rationale for assigning a larger rank to capture global patterns. The mid-level shared LoRA has a moderate effective rank, reflecting its role in modeling partially shared features across specific layers—intermediate in complexity between global and fine-grained representations—consistent with our expectations for mid-scale adaptation. In contrast, the layer-specific LoRA shows the lowest effective rank, suggesting that it primarily serves fine-grained adjustments at individual layers. The information captured is more targeted and contains less redundancy, confirming that low-rank configurations are sufficient for efficient adaptation at the layer level.

\begin{figure}[t]
\includegraphics[width=\textwidth]{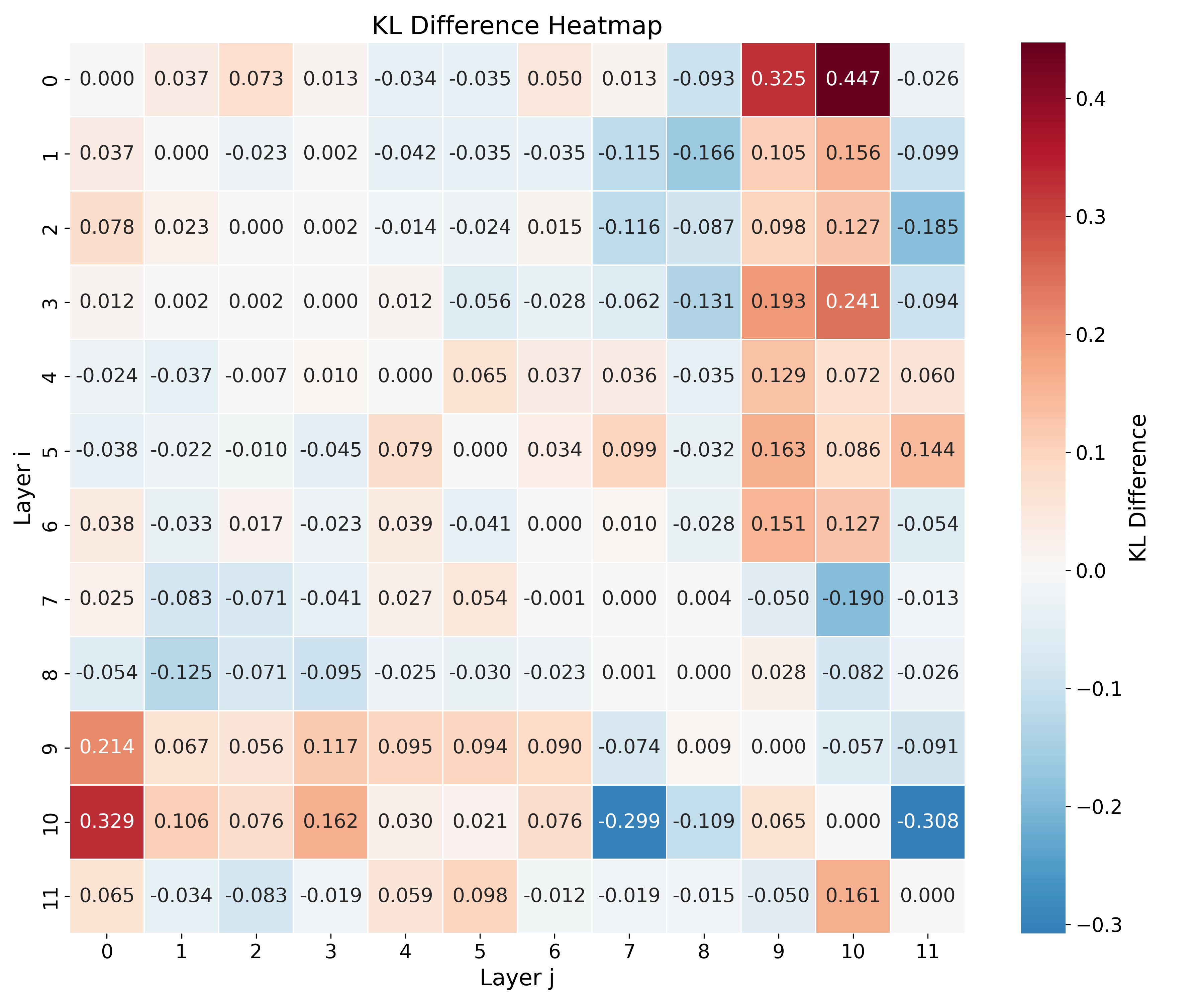}
\caption{KL divergence difference heatmap between MSPLoRA and standard LoRA, measuring the divergence of singular value spectra across layer pairs. Each cell \((i,j)\) indicates how much more (or less) MSPLoRA separates the layer-specific LoRA components than standard LoRA. Positive values—especially along the adjacent layer bands—suggest reduced redundancy and enhanced layer-specific modeling in MSPLoRA.} \label{fig4}
\end{figure}

\subsection{Analysis of Reducing Inter-Layer Information Redundancy}
Due to the high structural similarity of the Transformer architecture~\cite{transformer}, the hierarchical pattern inherently exhibits strong correlations, leading to adjacent LoRA components learning highly redundant information. MSPLoRA extracts shared information through global and mid-level LoRA, allowing layer-specific LoRA components to focus more on localized information, thereby reducing inter-layer redundancy.

To verify this, we conduct detailed experiments, utilizing divergence-based similarity metrics to measure the correlations between different LoRA components. The singular value spectrum similarity metric is used to quantify the differences in singular value distributions among different LoRA components, which can be expressed as:

\begin{equation}
    D_{\text{KL}}(\Sigma_i \parallel \Sigma_j) = \sum_k p_k^{(i)} \log \frac{p_k^{(i)}}{p_k^{(j)}}
\end{equation}

where

\begin{equation}
p_k^{(i)} = \frac{\sigma_k^{(i)}}{\sum_{m=1}^{M} \sigma_m^{(i)}}, \quad p_k^{(j)} = \frac{\sigma_k^{(j)}}{\sum_{m=1}^{M} \sigma_m^{(j)}}
\end{equation}

Here, \( \sigma_k^{(i)} \) and \( \sigma_k^{(j)} \) represent the k-th singular values of layer \( i \) and layer \( j \), respectively.

If the KL divergence is large, it indicates less redundancy in inter-layer information. Conversely, if the KL divergence is small, it suggests that different layers exhibit high similarity, implying a significant amount of shared redundant information.We fine-tune the RoBERTa-base model on MRPC, STS-B, and CoLA using both standard LoRA and MSPLoRA, and record the average SVD results of each layer’s LoRA components under both settings. We then compute the spectral similarity between each pair of layers based on their singular value distributions. The hyperparameter settings are consistent with those used in the experiments in Section 4. As shown in Fig 4, under the standard LoRA configuration, the LoRA components across different layers exhibit high similarity, indicating significant redundancy in the learned representations. In contrast, MSPLoRA leads to a notable reduction in similarity among the layer-specific LoRA components, suggesting that it effectively mitigates inter-layer redundancy and encourages each component to focus on layer-specific information.

\section{Conclusion}
This study proposes MSPLoRA, a parameter-efficient fine-tuning method based on a multi-scale pyramid structure. Traditional LoRA applies a fixed low-rank decomposition across all layers, failing to effectively model information at different hierarchical levels. To address this limitation, MSPLoRA introduces global shared LoRA, mid-level shared LoRA, and layer-specific LoRA, which respectively capture global patterns, mid-level features, and fine-grained information, thereby enhancing both model adaptability and parameter efficiency. We conduct comprehensive evaluations on natural language understanding tasks and instruction-following tasks. Experimental results demonstrate that MSPLoRA reduces the number of trainable parameters while maintaining or even surpassing the performance of standard LoRA and its variants, validating its generalization ability across different tasks and model scales. Additionally, ablation studies further analyze the contributions of different LoRA components, confirming the effectiveness of the multi-scale structure in information modeling. Overall, MSPLoRA exhibits superior performance in reducing computational costs, improving parameter efficiency, and enhancing hierarchical adaptation, offering a new optimization framework for the efficient fine-tuning of large-scale pre-trained models.

\bibliographystyle{splncs04}
\bibliography{ref}

\begin{thebibliography}{10}
\providecommand{\url}[1]{\texttt{#1}}
\providecommand{\urlprefix}{URL }
\providecommand{\doi}[1]{https://doi.org/#1}

\bibitem{gpt}
Achiam, J., Adler, S., Agarwal, S., Ahmad, L., Akkaya, I., Aleman, F.L., Almeida, D., Altenschmidt, J., Altman, S., Anadkat, S., et~al.: Gpt-4 technical report. arXiv preprint arXiv:2303.08774  (2023)

\bibitem{bhojanapalli}
Bhojanapalli, S., Chakrabarti, A., Veit, A., Lukasik, M., Jain, H., Liu, F., Chang, Y.W., Kumar, S.: Leveraging redundancy in attention with reuse transformers. arXiv preprint arXiv:2110.06821  (2021)

\bibitem{stsb}
Cer, D., Diab, M., Agirre, E., Lopez-Gazpio, I., Specia, L.: Semeval-2017 task 1: Semantic textual similarity-multilingual and cross-lingual focused evaluation. arXiv preprint arXiv:1708.00055  (2017)

\bibitem{heval}
Chen, M., Tworek, J., Jun, H., Yuan, Q., Pinto, H.P.D.O., Kaplan, J., Edwards, H., Burda, Y., Joseph, N., Brockman, G., et~al.: Evaluating large language models trained on code. arXiv preprint arXiv:2107.03374  (2021)

\bibitem{instructeval}
Chia, Y.K., Hong, P., Bing, L., Poria, S.: Instructeval: Towards holistic evaluation of instruction-tuned large language models. arXiv preprint arXiv:2306.04757  (2023)

\bibitem{dalvi}
Dalvi, F., Sajjad, H., Durrani, N., Belinkov, Y.: Analyzing redundancy in pretrained transformer models. arXiv preprint arXiv:2004.04010  (2020)

\bibitem{qlora}
Dettmers, T., Pagnoni, A., Holtzman, A., Zettlemoyer, L.: Qlora: Efficient finetuning of quantized llms. Advances in neural information processing systems  \textbf{36},  10088--10115 (2023)

\bibitem{bert}
Devlin, J., Chang, M.W., Lee, K., Toutanova, K.: Bert: Pre-training of deep bidirectional transformers for language understanding. In: Proceedings of the 2019 conference of the North American chapter of the association for computational linguistics: human language technologies, volume 1 (long and short papers). pp. 4171--4186 (2019)

\bibitem{ft}
Ding, N., Qin, Y., Yang, G., Wei, F., Yang, Z., Su, Y., Hu, S., Chen, Y., Chan, C.M., Chen, W., et~al.: Parameter-efficient fine-tuning of large-scale pre-trained language models. Nature Machine Intelligence  \textbf{5}(3),  220--235 (2023)

\bibitem{mrpc}
Dolan, B., Brockett, C.: Automatically constructing a corpus of sentential paraphrases. In: Third international workshop on paraphrasing (IWP2005) (2005)

\bibitem{drop}
Dua, D., Wang, Y., Dasigi, P., Stanovsky, G., Singh, S., Gardner, M.: Drop: A reading comprehension benchmark requiring discrete reasoning over paragraphs. arXiv preprint arXiv:1903.00161  (2019)

\bibitem{clinical}
Gema, A.P., Minervini, P., Daines, L., Hope, T., Alex, B.: Parameter-efficient fine-tuning of llama for the clinical domain. arXiv preprint arXiv:2307.03042  (2023)

\bibitem{sppnet}
He, K., Zhang, X., Ren, S., Sun, J.: Spatial pyramid pooling in deep convolutional networks for visual recognition. IEEE transactions on pattern analysis and machine intelligence  \textbf{37}(9),  1904--1916 (2015)

\bibitem{mmlu}
Hendrycks, D., Burns, C., Basart, S., Zou, A., Mazeika, M., Song, D., Steinhardt, J.: Measuring massive multitask language understanding. arXiv preprint arXiv:2009.03300  (2020)

\bibitem{adapter1}
Houlsby, N., Giurgiu, A., Jastrzebski, S., Morrone, B., De~Laroussilhe, Q., Gesmundo, A., Attariyan, M., Gelly, S.: Parameter-efficient transfer learning for nlp. In: International conference on machine learning. pp. 2790--2799. PMLR (2019)

\bibitem{lora(adapter2)}
Hu, E.J., Shen, Y., Wallis, P., Allen-Zhu, Z., Li, Y., Wang, S., Wang, L., Chen, W., et~al.: Lora: Low-rank adaptation of large language models. ICLR  \textbf{1}(2), ~3 (2022)

\bibitem{aslora}
Hu, J., Xiao, X., Zhang, M., Chen, Y., Ren, Z., Chen, Z., Ren, P.: Aslora: Adaptive sharing low-rank adaptation across layers. arXiv preprint arXiv:2412.10135  (2024)

\bibitem{prefixtuning}
Li, X.L., Liang, P.: Prefix-tuning: Optimizing continuous prompts for generation. arXiv preprint arXiv:2101.00190  (2021)

\bibitem{fpn}
Lin, T.Y., Doll{\'a}r, P., Girshick, R., He, K., Hariharan, B., Belongie, S.: Feature pyramid networks for object detection. In: Proceedings of the IEEE conference on computer vision and pattern recognition. pp. 2117--2125 (2017)

\bibitem{codepeft}
Liu, S., Keung, J., Yang, Z., Liu, F., Zhou, Q., Liao, Y.: Delving into parameter-efficient fine-tuning in code change learning: An empirical study. In: 2024 IEEE International Conference on Software Analysis, Evolution and Reengineering (SANER). pp. 465--476. IEEE (2024)

\bibitem{ptuning1}
Liu, X., Ji, K., Fu, Y., Tam, W., Du, Z., Yang, Z., Tang, J.: {P}-tuning: Prompt tuning can be comparable to fine-tuning across scales and tasks. In: Muresan, S., Nakov, P., Villavicencio, A. (eds.) Proceedings of the 60th Annual Meeting of the Association for Computational Linguistics (Volume 2: Short Papers). pp. 61--68. Association for Computational Linguistics, Dublin, Ireland (May 2022). \doi{10.18653/v1/2022.acl-short.8}, \url{https://aclanthology.org/2022.acl-short.8/}

\bibitem{ptuning2}
Liu, X., Zheng, Y., Du, Z., Ding, M., Qian, Y., Yang, Z., Tang, J.: Gpt understands, too. AI Open  \textbf{5},  208--215 (2024)

\bibitem{roberta}
Liu, Y., Ott, M., Goyal, N., Du, J., Joshi, M., Chen, D., Levy, O., Lewis, M., Zettlemoyer, L., Stoyanov, V.: Roberta: A robustly optimized bert pretraining approach. arXiv preprint arXiv:1907.11692  (2019)

\bibitem{pissa}
Meng, F., Wang, Z., Zhang, M.: Pissa: Principal singular values and singular vectors adaptation of large language models. Advances in Neural Information Processing Systems  \textbf{37},  121038--121072 (2024)

\bibitem{nadipalli}
Nadipalli, S.: Layer-wise evolution of representations in fine-tuned transformers: Insights from sparse autoencoders. arXiv preprint arXiv:2502.16722  (2025)

\bibitem{financial}
Olariu, I., Lothritz, C., Klein, J., Bissyand{\'e}, T., Guo, S., Haddadan, S.: Evaluating parameter-efficient finetuning approaches for pre-trained models on the financial domain. In: Bouamor, H., Pino, J., Bali, K. (eds.) Findings of the Association for Computational Linguistics: EMNLP 2023. pp. 15482--15491. Association for Computational Linguistics, Singapore (Dec 2023). \doi{10.18653/v1/2023.findings-emnlp.1035}, \url{https://aclanthology.org/2023.findings-emnlp.1035/}

\bibitem{t5}
Raffel, C., Shazeer, N., Roberts, A., Lee, K., Narang, S., Matena, M., Zhou, Y., Li, W., Liu, P.J.: Exploring the limits of transfer learning with a unified text-to-text transformer. Journal of machine learning research  \textbf{21}(140),  1--67 (2020)

\bibitem{qnli}
Rajpurkar, P., Zhang, J., Lopyrev, K., Liang, P.: Squad: 100,000+ questions for machine comprehension of text. arXiv preprint arXiv:1606.05250  (2016)

\bibitem{adapterd}
R{\"u}ckl{\'e}, A., Geigle, G., Glockner, M., Beck, T., Pfeiffer, J., Reimers, N., Gurevych, I.: Adapterdrop: On the efficiency of adapters in transformers. arXiv preprint arXiv:2010.11918  (2020)

\bibitem{sst2}
Socher, R., Perelygin, A., Wu, J., Chuang, J., Manning, C.D., Ng, A.Y., Potts, C.: Recursive deep models for semantic compositionality over a sentiment treebank. In: Proceedings of the 2013 conference on empirical methods in natural language processing. pp. 1631--1642 (2013)

\bibitem{bbh}
Srivastava, A., Rastogi, A., Rao, A., Shoeb, A.A.M., Abid, A., Fisch, A., Brown, A.R., Santoro, A., Gupta, A., Garriga-Alonso, A., et~al.: Beyond the imitation game: Quantifying and extrapolating the capabilities of language models. arXiv preprint arXiv:2206.04615  (2022)

\bibitem{hrnet}
Sun, K., Xiao, B., Liu, D., Wang, J.: Deep high-resolution representation learning for human pose estimation. In: Proceedings of the IEEE/CVF conference on computer vision and pattern recognition. pp. 5693--5703 (2019)

\bibitem{alpaca}
Taori, R., Gulrajani, I., Zhang, T., Dubois, Y., Li, X., Guestrin, C., Liang, P., Hashimoto, T.B.: Stanford alpaca: An instruction-following llama model (2023)

\bibitem{llama}
Touvron, H., Lavril, T., Izacard, G., Martinet, X., Lachaux, M.A., Lacroix, T., Rozi{\`e}re, B., Goyal, N., Hambro, E., Azhar, F., et~al.: Llama: Open and efficient foundation language models. arXiv preprint arXiv:2302.13971  (2023)

\bibitem{llama2}
Touvron, H., Martin, L., Stone, K., Albert, P., Almahairi, A., Babaei, Y., Bashlykov, N., Batra, S., Bhargava, P., Bhosale, S., et~al.: Llama 2: Open foundation and fine-tuned chat models. arXiv preprint arXiv:2307.09288  (2023)

\bibitem{dylora}
Valipour, M., Rezagholizadeh, M., Kobyzev, I., Ghodsi, A.: Dylora: Parameter efficient tuning of pre-trained models using dynamic search-free low-rank adaptation. arXiv preprint arXiv:2210.07558  (2022)

\bibitem{transformer}
Vaswani, A., Shazeer, N., Parmar, N., Uszkoreit, J., Jones, L., Gomez, A.N., Kaiser, {\L}., Polosukhin, I.: Attention is all you need. Advances in neural information processing systems  \textbf{30} (2017)

\bibitem{glue}
Wang, A., Singh, A., Michael, J., Hill, F., Levy, O., Bowman, S.R.: Glue: A multi-task benchmark and analysis platform for natural language understanding. arXiv preprint arXiv:1804.07461  (2018)

\bibitem{vlm}
Wang, W., Chen, Z., Chen, X., Wu, J., Zhu, X., Zeng, G., Luo, P., Lu, T., Zhou, J., Qiao, Y., et~al.: Visionllm: Large language model is also an open-ended decoder for vision-centric tasks. Advances in Neural Information Processing Systems  \textbf{36},  61501--61513 (2023)

\bibitem{cola}
Warstadt, A., Singh, A., Bowman, S.R.: Neural network acceptability judgments. Transactions of the Association for Computational Linguistics  \textbf{7},  625--641 (2019)

\bibitem{code}
Yang, Z., Liu, F., Yu, Z., Keung, J.W., Li, J., Liu, S., Hong, Y., Ma, X., Jin, Z., Li, G.: Exploring and unleashing the power of large language models in automated code translation. Proceedings of the ACM on Software Engineering  \textbf{1}(FSE),  1585--1608 (2024)

\bibitem{adalora(onlyqv)}
Zhang, Q., Chen, M., Bukharin, A., Karampatziakis, N., He, P., Cheng, Y., Chen, W., Zhao, T.: Adalora: Adaptive budget allocation for parameter-efficient fine-tuning. arXiv preprint arXiv:2303.10512  (2023)

\bibitem{lily}
Zhong, Y., Zhou, Y.: Low-rank interconnected adaptation across layers. arXiv preprint arXiv:2407.09946  (2024)

\bibitem{detr}
Zhu, X., Su, W., Lu, L., Li, B., Wang, X., Dai, J.: Deformable detr: Deformable transformers for end-to-end object detection. arXiv preprint arXiv:2010.04159  (2020)

\end{thebibliography}

\end{document}